\documentclass{article}
\usepackage{ijcai18}

\usepackage{times}
\usepackage{xcolor}
\usepackage{soul}
\usepackage[utf8]{inputenc}
\usepackage[small]{caption}

\usepackage[vlined,ruled]{algorithm2e}
\usepackage{amsmath}
\usepackage{amssymb}
\usepackage{amsthm}
\usepackage{array}
\usepackage{color}
\usepackage{courier}
\usepackage{dsfont}
\usepackage{graphicx}
\usepackage{helvet}
\usepackage{mathtools}
\usepackage{multicol}
\usepackage{pgfplots}
\usepackage[pdftex]{pstricks}
\usepackage{tikz}
\usepackage{url}
\usepackage{verbatim}
\usepackage{nameref}
\usetikzlibrary{shapes,fit}

\usepackage{natbib}

\title{
	Open Loop Execution of Tree-Search Algorithms\\
	\vspace{5mm}
	Extended version
	\vspace{5mm}
}

\author{
	Erwan Lecarpentier$^{1,2}$,
	Guillaume Infantes$^1$,
	Charles Lesire$^1$,
	Emmanuel Rachelson$^2$\\
	$^1$ ONERA - The French Aerospace Lab,\\
	2 avenue Edouard Belin, 31055 Toulouse, France\\
	$^2$ ISAE - SUPAERO,
	University of Toulouse,\\
	10 avenue Edouard Belin, 31055 Toulouse, France\\
	$^1$ \{first name\}.\{last name\}@onera.fr,
	$^2$ \{first name\}.\{last name\}@isae-supaero.fr
}
\newtheorem{lemma}{Lemma}
\newtheorem{thm}{Theorem}

\def\addproofs{1} 

\def\commentcolor{blue!60!black} 
\def\algocomment#1{\texttt{\footnotesize \color{\commentcolor}// #1}}

\def\wrt{w.r.t.}
\def\ie{i.e.}
\def\eg{e.g.}

\DeclareMathOperator*{\argmax}{arg\,max}

\def\intrange#1#2{\ensuremath{\left\{#1, \cdots, #2\right\}}} 
\def\return{X} 
\def\estimate{\overline{\return}} 

\def\algoname{OLTA} 
\def\algonamemeaning{Open Loop Tree-search Algorithm} 
\def\oluct{OLUCT}
\def\oluctmeaning{Open Loop UCT}
\def\ucb{UCB}
\def\ucbmeaning{Upper Confidence Bound}

\def\returnestimate#1#2#3{
	\ensuremath {
		\estimate^{#1}_{#2,#3}
	}
}

\def\nbsamples#1#2#3{
	\ensuremath {
		T^{#1}_{#2,#3}
	}
}

\begin{document}

\maketitle

\begin{abstract}
In the context of tree-search stochastic planning algorithms where a generative model is available, we consider on-line planning algorithms building trees in order to recommend an action. We investigate the question of avoiding re-planning in subsequent decision steps by directly using sub-trees as action recommender. Firstly, we propose a method for open loop control via a new algorithm taking the decision of re-planning or not at each time step based on an analysis of the statistics of the sub-tree. Secondly, we show that the probability of selecting a suboptimal action at any depth of the tree can be upper bounded and converges towards zero. Moreover, this upper bound decays in a logarithmic way between subsequent depths. This leads to a distinction between node-wise optimality and state-wise optimality. Finally, we empirically demonstrate that our method achieves a compromise between loss of performance and computational gain.
\end{abstract}

\section{Introduction}

Tree-search based algorithms recently encountered a real success at solving sequential, highly combinatorial problems such as the challenging game of Go \citep{enzenberger2010fuego,silver2016mastering}.
Such algorithms use a generative model of the environment to simulate episodes starting from the current state of the agent \citep{sutton1991dyna,sutton1998reinforcement}. This allows the exploration of reachable states and actions and results in the construction of an (unbalanced) scenario tree, that aims at identifying promising branches with a limited computational budget. When the computational budget is exhausted, the recommended action at the root node is applied and a new tree is built in the resulting state.
This results overall in a closed loop control process.

%
We are interested in stochastic problems with large state spaces (\eg{} continuous) with a short decision time (budget).
%
In this setting, open loop planning algorithms have proven to be successful~\citep{bubeck2010open} and even to outperform~\citep{weinstein2012bandit} the standard approaches that consider closed loop policy trees such as UCT \citep{kocsis2006bandit}.
They seek for optimal sequences of actions (plans) rather than optimal policies despite the sub-optimal nature of a plan in stochastic environments.
Indeed, computing the latter prevents feed-back on the explored states but allows to break the complexity of the state space exploration.
%
Given a tree computed by an open loop planning algorithm, we propose to keep the sub-tree reached by the application of the recommended action and to directly use it as the main tree for the subsequent time step, without re-planning.
What motivates this approach is sparing the computational cost of tree building for subsequent time steps, hence reducing the number of calls to the simulator.
The interest of this can be seen in two ways.
On one hand it is a way of reducing energy consumption for systems with low computational resources \citep{wilson2014bounded,wilson2016toward}.
On the other hand, the saved computational time can be re-invested into other tasks.
Particularly, this approach is adapted for low level control (\ie{} high frequency) where sub-sequent tree developments is cumbersome.
%
In this framework, \citet{perez2012monte} and \citet{heusner2011uct} considered keeping the tree in deterministic environments but observed a negative impact as the sub-trees were systematically kept without analysis.
Moreover, they lose the aforementioned computational gain by refining the sub-trees.

In this paper, we study the impact of using the subsequent sub-trees as main trees for the next action steps without further re-planning.
We claim that in lowly-stochastic environments, the reached performance is comparable to algorithms systematically discarding the tree.
Our contribution is threefold.
(1) We introduce a new algorithm called \algoname{} (Section~\ref{sec:algo}), performing a systematic analysis of the sub-tree and taking the decision of re-planning or not at each time step.
(2) We upper bound the probability of selecting a suboptimal action within a sub-tree, the sense of optimality being defined in an open loop fashion (Section~\ref{sec:theory}).
Additionally, we show that this upper bound decays logarithmically with the sub-tree depth.
(3) We show in our experiments the benefit of applying such a method both in terms of performance and computational cost saving (Section~\ref{sec:experiments}).



\section{Background}
\label{sec:background}

\subsection{Markov Decision Process}
We model the planning problem as a Markov Decision Process (MDP) where an agent sequentially takes actions with the general goal of maximizing the cumulative return fed back by the environment \citep{puterman2014markov}.
We refer to the state space as $S$ and the action space as $A$.
We suppose the number of actions to be finite with $K = |A|$, thus we write $A = \{ a_i \}_{i=1}^K$.
We also consider that the available actions are independent of the state the agent lies in.
The state transition function is stochastic and we note $P(s' | s, a)$ the probability of reaching state $s'$ after taking action $a$ in state $s$.
The reward model is denoted by $r(s,a,s')$ and refers to the scalar reward received while performing the transition $(s,a,s')$.
We assume that this reward function is deterministic.
Finally, we suppose the horizon of the MDP is infinite and we note $\gamma \in [0,1)$ the discount factor which represents the importance of the subsequent collected rewards.

\subsection{Tree Representation}
\label{sec:background:tree}
When a generative model of the MDP is available, it becomes possible to use it within planning algorithms.
Tree-search algorithms use this model in order to build a tree of what may possibly occur in the current situation of the agent \citep{sutton1991dyna,sutton1998reinforcement,silver2008sample}.
In the stochastic setting with potentially infinitely many states, we use a tree structure similar to the one used by \citet{bubeck2010open}.
The tree built at each time step consists in a look-ahead search of the possible outcomes while following some action plan starting from the current state of the agent $s_0 \in S$.
Thus, the root node of the tree is labelled by the unique state $s_0$.
The edges correspond to the $K$ available actions, $K$ being the branching factor of the tree.
The tree itself conforms to an ensemble of action sequences, or plans, originating from its root node.

We emphasize the fact that this tree structure implies that we search for a state-independent optimal sequence of actions (open loop plan) which is in general sub-optimal compared to a state-dependent policy search.
The THTS family of algorithms in particular \citep{keller2013trial} defines trees with chance and decision nodes while our structure does not apply an equality operator on the sampled states.
Following \citet{bubeck2010open,weinstein2012bandit}, we argue that closed-loop application of the first action in optimal open loop plans, although theoretically suboptimal, can be competitive with these methods in practice, while being more sample-efficient.

\begin{figure}
	\centering
	\includegraphics[
		trim={2cm 0cm 2cm 0cm},
		clip,
		height=4.5cm
	]{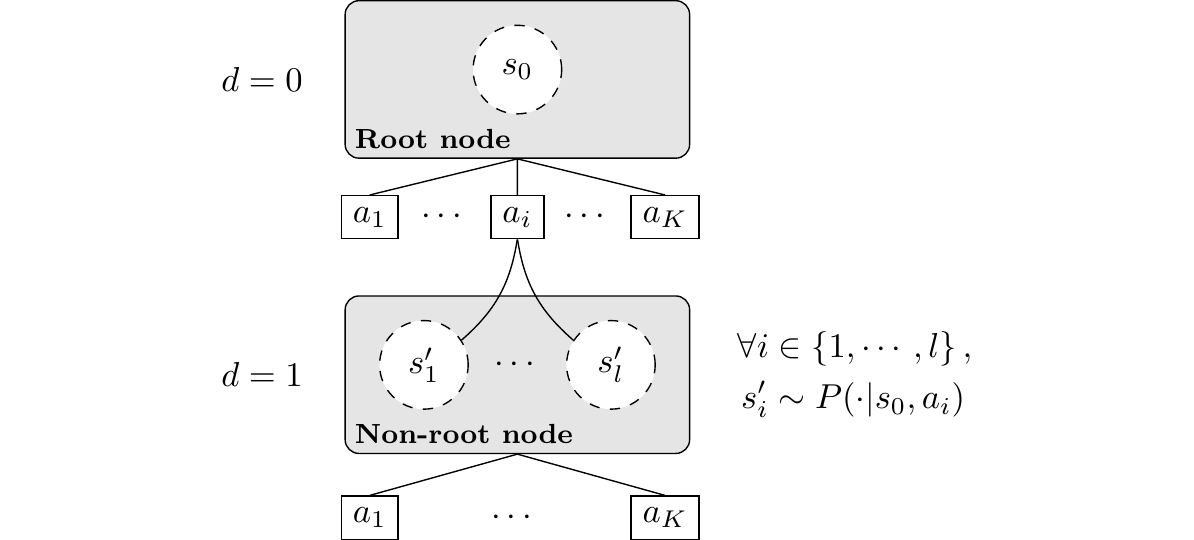}
	\caption{
		General representation of a tree, where $l \in \mathbb{N}$ is the number of times the sub-tree reached by action $a_i$ has been developed. Two nodes are represented in this tree with their respective depths on the left.
	}
	\label{fig:tree}
\end{figure}

Since the transition model is stochastic, the non-root nodes are not labelled by a unique state.
Instead, every such node is associated to a state distribution resulting from the application of the action plan leading to the considered node and starting from $s_0$.
During the exploration, we consider saving all sampled states at each non-root node.
A comprehensive illustration of such a tree can be found in Figure~\ref{fig:tree}.
This approach extends straightforwardly to Partially Observable Markov Decision Processes (POMDP) \citep{silver2010monte}.

Given a tree-search, open loop planning algorithm, we call $\mathcal{T}_d$ the \textbf{tree at depth $\boldsymbol d \in \mathbb{N}$}, that is the sub-tree resulting from the application of the $d$ first recommended actions.
Hence $\mathcal{T}_0$ denotes the whole tree, $\mathcal{T}_1$ the tree starting from the node reached by the application of the first recommended action and so on.

\subsection{\oluctmeaning{}}

For the sake of clarity and in order to clearly separate the tree building properties from the open loop execution presented in the next section, we define an open loop planning algorithm utilizing the presented tree structure that we call \oluctmeaning{} (\oluct{}).
The difference between UCT and \oluct{} is that \oluct{} is not provided with an equality operator over states.
Within the THTS terminology, this means that decision and chance nodes do not correspond to a single state but to the state distribution reachable by the action plan leading to the node.
Hence decision and chance nodes are associated to the state distribution which makes \oluct{} an open loop planning algorithm.
The fundamental consequence is that an action value within our tree is computed \wrt{} the parent node's state distribution rather than a single state.

Apart from this, \oluct{} uses the same exploration procedure as UCT.
Within a node, we note $\estimate_{i,u}$ the estimated expected return of action $i$ after $u$ samples of this action.
$T_i(t)$ is the number of trials of action $i$ up to time $t$ of the \oluct{} procedure.
An \ucbmeaning{} (\ucb{}) strategy~\citep{auer2002finite} is applied at each node where each action is seen as an arm of a bandit problem.
The tree policy selects the action $I_t$ with the highest \ucb{}:
\begin{equation*}
	I_t = \argmax_{i \in \intrange{1}{K}} \left\{ \estimate_{i,T_i(t-1)} + c_{t-1,T_i(t-1)} \right\}
	,
\end{equation*}
where $c_{t,u} = 2 C_p \sqrt{\frac{ln(t)}{u}}$ is an exploration term ensuring that all actions will be sampled infinitely often.
The $C_p$ parameter  drives the exploration-exploitation trade-off.
The \oluct{} tree building procedure is detailed in Algorithm~\ref{alg:oluct}.

\begin{algorithm}
	\DontPrintSemicolon
	\textbf{Function} createTree(state $s$):\;
	\textbf{Parameters:} budget $n$; default policy $\pi_{default}$.\;
	Create root node $\nu_{root}(s)$\;
	\For{t $\in \{1, \cdots, n\}$} {
		$\nu_{leaf} =$ Select($\nu_{root}$);
		\algocomment{Select a leaf node \wrt{} the UCT strategy and sample a new state for each encountered node.}\;
		
		Expand($\nu_{leaf}$);
		\algocomment{Expand the node if not terminal using the generative model.}\;
		
		$\Delta =$ Evaluate($\nu_{leaf}, \pi_{default}$);
		\algocomment{Simulate a roll-out using $\pi_{default}$, starting from the last sampled state in $\nu_{leaf}$.}\;
		
		Backup($\nu_{leaf}, \Delta$);
		\algocomment{Back-propagate the sampled return.}\;
	}
	\textbf{return} $\mathcal{T}(\nu_{root})$
	\caption{\oluct{} tree building procedure}
	\label{alg:oluct}
\end{algorithm}

\section{\algoname{} (\algonamemeaning{})}
\label{sec:algo}

\subsection{Description}
In order to control the execution of open loop plans, 
we propose a new algorithm called \algoname{} (Algorithm \ref{alg:mainalgo}).
It relies on a generic open loop planning algorithm to generate a tree, rooting from the current state.
For the next execution time step, it decides either to use the sub-tree reached by the recommended action or to trigger a re-planning by building a new tree.
If no re-planning is triggered, then the recommended action of the sub-tree is applied without using the additional information of the new state observed
after the transition.
This results in an open loop control process and spares the cost of developing a new tree starting at this state.
The intuition behind \algoname{} is that several consecutive recommended actions in an optimal branch of the tree can be reliable, despite the randomness of the environment.
A major example of such a case is low-level control, where consecutive sampled states are close to each other.

%
In this paper, for its performance and simplicity, we chose to implement \oluct{} as the open loop planning algorithm utilized by \algoname{}.
However, any other algorithm generating trees as described in Section~\ref{sec:background:tree} could be used in the same way (\eg{} OLOP~\citep{bubeck2010open}, or HOLOP~\citep{weinstein2012bandit}).

One important feature of \algoname{} is the so-called ``\emph{decisionCriterion}'', based on which the agent decides to either use the first sub-tree following the recommended action, or to re-build a new tree from the current state.
The decision is based on a comparison with the characteristics of the resulting sub-tree and the current state of the agent.
In the next section, we discuss different decision criteria, leading to the consideration of a family of different algorithms.


\begin{algorithm}
	\DontPrintSemicolon
	\textbf{Function} \algoname{}:\;
	\textbf{Parameters:}
	initial state $s_0$;
	tree building procedure $\textrm{createTree}$;
	re-planning criterion $\textrm{decisionCriterion}$.\;
	$s = s_0$;\;
	$\mathcal{T} = \textrm{createTree}(s)$;\;
	\While{s is not terminal}{
		\If{decisionCriterion($s,\mathcal{T}$)}{
			$a$ = recommendedAction($\mathcal{T}$);
			\algocomment{Get the first recommended action.}\;
		}
		\Else{
			$\mathcal{T} = \textrm{createTree}(s)$;
			\algocomment{Create a new tree from the current state.}\;
			$a$ = recommendedAction($\mathcal{T}$);
			\algocomment{Get the first recommended action.}\;
		}
		$\mathcal{T} =$ subTree($\mathcal{T}, a$);
		\algocomment{Move the tree to the first sub-tree resulting from the application of $a$.}\;
		s := realWorldTransitionFunction($s,a$);
	}
	\caption{\algoname{} algorithm}
	\label{alg:mainalgo}
\end{algorithm}


\subsection{Decision Criterion}
\label{sec:algo:decision_criterion}
The simplest implementation of the decision criterion is to keep the sub-tree only if its root node is fully expanded.
This means that each action has been sampled at least once.
We call the resulting algorithm \textbf{Plain \algoname{}}.
It naively trusts the value estimates of the sub-tree, thus applies the whole plan of recommended actions at each depth until it reaches a partially expanded node.
Therefore, Plain \algoname{} is expected to perform better in deterministic environments.
In stochastic cases however, those estimates may be biased because of the different sources of uncertainty within the MDP (reward function, state transition function and action selection).
For this reason, we seek more robust criteria to base the decision on.

A natural way to decide whether to keep the sub-tree or not is to track if the recommended action is optimal \wrt{} the new state $s$ of the agent.
Here we make an important distinction between a \textbf{state-wise optimal action} and a \textbf{node-wise optimal action}.
The first one is the action recommended by the optimal policy in a specific state.
We note it $a^* = \argmax_{a \in A} Q^*(s,a)$, with $Q^*:S \times A \rightarrow \mathbb{R}$ the optimal state-action value function.
In order to define the second one, we introduce $S_d$, the state random variable at the root node of $\mathcal{T}_d$.
Its distribution results from the application of the $d$ first recommended actions starting from $s_0$, so $S_d \sim P( \cdot | s_0, a_0, \cdots, a_{d-1} ) \equiv P_{S_d}(\cdot)$.
The node-wise optimal action maximizes the expected return given the state \emph{distribution} of the node.
We note it $a_d^* = \argmax_{a \in A} Q^*_d(a)$ where $Q^*_d: A \rightarrow \mathbb{R}$ is the optimal action value function \wrt{} the state distribution at the root node of $\mathcal{T}_d$, that is $Q^*_d(a) = \mathbb{E}_{s\sim P_{S_d}} (Q^*(s,a))$.
Following \citet{bellemare2017distributional}, a distributional Bellman equation can be expressed in terms of three sources of randomness that are: $R: S \times A \rightarrow \mathbb{R}$ the stochastic reward function; $\return: S \times A \rightarrow \mathbb{R}$ the random return; and $P^\pi$ the transition operator with $P^\pi \return(s,a) \stackrel{D}{\coloneqq} \return(S',A')$, $S' \sim P( \cdot | s,a)$ and $A' \sim \pi(\cdot | S')$.
Mathematically, we have the following distributional Bellman equations:
\begin{equation*}
\begin{cases}
Q^\pi(s,a) = \mathbb{E}_\pi (\return(s,a))\\
Q^\pi_d(a) = \mathbb{E}_{P_{S_d}} (Q^\pi(s,a))
\end{cases}
,
\end{equation*}
with $\return(s,a) \sim R(s,a) + \gamma P^\pi \return(s,a)$ and $S_d \sim P_{S_d}(\cdot)$.
Unfortunately, at the root node of $\mathcal{T}_d$ for $d > 0$, open loop tree-search algorithms do not estimate $Q^*$ but $Q_d^*$.
The bias introduced by the state distribution implies that in the general case we have no guarantee that $a^* = a_d^*$.
The risk is that the set $\Omega_{S_d}$ of possible realizations of $S_d$ can include states where $a^*$ is sub-optimal, in which case the resulting return evaluations would weight in favour of a different action than $a^*$.
In other words --- introducing the notion of domination domain for an action $a$ as $\mathcal{D}_a = \{ s \in S | \pi^*(s) = a \} \subset S$ --- if $\Omega_{S_d}$ is not included in $\mathcal{D}_{a^*}$, then the risk of the recommended action to be state-wise sub-optimal is increased.
Conversely, if $\Omega_{S_d}$ is included in the domination domain of $a^*$, then the optimal action will be selected given that the budget is ``big enough'' \wrt{} the chosen tree-search algorithm's performance. 
Consequently, one should base the decision criterion on the analysis of $P_{S_d}$ and the action domination domains.
To compute these domains, \citet{rachelson2010locality} use the properties of Lipschitz-MDPs.
Although the following discussion is inspired by this work, the consideration of Lipschitz-MDPs is out of the scope of this paper.
We discuss below the construction of decision criteria that will be illustrated in Section~\ref{sec:experiments}.

\textbf{Current state analysis \& POMDP setting.}
The current state $s$ of the agent can be compared to the empirical state distribution $\overline{P_{S_d}}$ at the root node of the sub-tree.
If $P_{S_d}(s)$ is large, then the value estimators are related to the locality of the state space the agent lies in.
If not, then the node-wise optimal action may not be state-wise optimal.
This consideration supposes to identify a state-metric for which two close states have a high chance to be in the same action domination domain.
Alternatively, in the case of a POMDP, a belief distribution on the current state is available instead of the current state itself \citep{kaelbling1998planning}.
In such a case, a direct comparison between this distribution and $\overline{P_{S_d}}$ can be performed (\eg{} with a Wasserstein metric).
Note that making use of the current state of the agent makes the algorithm closed-loop, by definition.
We use the terminology "open-loop" in order to distinguish \algoname{} from classical closed-loop Tree Search algorithms that systematically re-plan, rooting from the current state (\eg{} OLOP \citep{bubeck2010open}, performs closed-loop execution).

\textbf{State distribution analysis.}
The dispersion and multi-modality of $\overline{P_{S_d}}$ could motivate not to re-use a sub-tree.
A high dispersion involves the possibility that $\overline{\Omega_{S_d}}$ does not belong to a single action domination domain and a re-planning should be triggered.
The same consideration applies in terms of multi-modality.
Conversely, a narrow, mono-modal, state distribution is a good hint for $\Omega_{S_d}$ to be comprised into a single action domination domain.

\textbf{Return distribution analysis.}
A widespread or a multi-modal return distribution for the recommended action in a node may indicate a strong dependency on the region of the state space we lie in.
If $\Omega_{S_d}$ covers different action domination domains, each of these domains may contribute a different return distribution to the node's return estimates, thus inducing a high variance on this distribution or even a multi-modality.
In this case, it could be beneficial to trigger the re-planning.
Alternatively, even after re-planning, widespread or multi-modal return distributions can naturally arise as a result of the MDP's reward and transition models.

We do not provide a unique generic method to base the decision criterion on.
Indeed, we believe that it is a strongly problem-dependent issue and that efficient heuristics can be built accordingly.
However, the analysis of the state and return distributions constitute promising indicators and we exemplify their use in the experiments of the last section.

\section{Theoretical Analysis}
\label{sec:theory}

In this section, we demonstrate that the algorithm asymptotically provides node-wise optimal actions for any sub-tree $\mathcal{T}_d$ of depth $d$.
We first derive an upper bound on the failure probability that converges towards zero when the initial budget $n$ of the algorithm goes to infinity.
Then, we characterize the loss of performance guarantees between subsequent depths and show a logarithmic decay of the upper bound.
The demonstration unfolds as follows: first we write a lower bound for the number of trials of the actions at the root of $\mathcal{T}_d$ in Lemma~\ref{lem:lemma1}; then we write an upper bound on the failure probability given a known budget at depth $d$ in Lemma~\ref{lem:lemma2}; finally we derive a recursive relation between the upper bounds of subsequent trees that leads to our result in Theorem~\ref{thm:thm1}.

We note $b(d) \in \mathbb{N}$ the \textbf{budget} used to develop $\mathcal{T}_d$ \ie{} the number of times the $d$ first recommended actions have been selected by the tree policy.
We note $\nbsamples{d}{i}{t}$ the number of times the $i^{\mathrm{th}}$ action at the root node of $\mathcal{T}_d$ has been selected by the \oluct{} tree policy after $t$ expansions of $\mathcal{T}_d$.
Similarly, $\estimate^d_{i,\nbsamples{d}{i}{t}} \equiv \estimate^d_{i,t}$ denotes the estimate of the return of the $i^{\mathrm{th}}$ action at depth $d$ after $t$ expansions of the sub-tree $\mathcal{T}_d$.
We write $I^d_t$ the index of the action chosen by the tree policy at depth $d$ after $t$ expansions of $\mathcal{T}_d$.
We have:
\begin{equation*}
	I^d_t = \argmax_{i \in \intrange{1}{K}}
	\left\{
	\returnestimate{d}{i}{t-1} +
	c_{t-1, \nbsamples{d}{i}{t-1}}
	\right\}
	.
\end{equation*}
The recommended action at depth $d$ given a budget $b(d)$ is $
	\widehat{I}^d = \argmax_{i \in \intrange{1}{K}}
	\returnestimate{d}{i}{b(d)}
$.
Following~\citet{kocsis2006bandit}, we assume that the empirical estimates $\returnestimate{d}{i}{t}$ converge and write $\return^d_{i,t} = \mathbb{E} \{ \returnestimate{d}{i}{t} \}$ and $\return^d_i = \lim_{t \rightarrow \infty} \return^d_{i,t}$.
Then, we define for $i \in \intrange{1}{K} \setminus i^*_d$, $\Delta_i^d = \return_{i^*_d}^d - \return_i^d $ where we note $i^*_d$ the index of the node-wise optimal action at the root node of $\mathcal{T}_d$.
We make the assumption that only one action is optimal in a given node.
The minimum return difference between a suboptimal action and the optimal one at depth $d$ is $\delta^d = \min_{i \in \intrange{1}{K} \setminus i^*_d} ( \Delta_i^d )$.

\begin{lemma}
	\label{lem:lemma1}
	\textbf{Lower bound for the number of trials.}
	For any sub-tree $\mathcal{T}_d$ developed with a budget $b(d)>K$, there exist a constant $\rho \ge 0$ such that $\nbsamples{d}{i}{b(d)} \ge \lceil \rho \ln(b(d)) \rceil$ for all $i \in \intrange{1}{K}$.
	Furthermore, we have the following sequence of lower bounds for the budget with $\lceil \cdot \rceil$ the ceiling function:
	\begin{equation*}
		\begin{cases}
			b(d=0) = n\\
			b(d) \ge \lceil \rho \ln(b(d-1)) \rceil 
		\end{cases}
		.
	\end{equation*}
\end{lemma}

\if\addproofs1
\begin{proof}
	The first result is borrowed from \citet{kocsis2006bandit} where they show it for a generic bandit problem.
	The extension to our case with a given budget is straightforward.
	The sequence of lower bounds can be derived by observing that $b(d) = \nbsamples{d-1}{\widehat{I}^{d-1}}{b(d-1)}$ and applying the previous lower bound.
\end{proof}
\fi

\begin{lemma}
	\label{lem:lemma2}
	\textbf{Upper bound on the failure probability at depth $d$ given the budget $b(d)$.}
	For any sub-tree $\mathcal{T}_d$ developed with a budget $b(d) > K$ we have the following upper bound on the failure probability, conditioned by the budget $b(d)$:
	\begin{equation*}
	P(\widehat{I}^d \neq i^*_d | b(d)) \le b(d)^{-\frac{\rho}{2} (\delta^d)^2} 
	.
	\end{equation*}
\end{lemma}

\if\addproofs1
\begin{proof}
	Let us first bound the failure probability with the probability of overestimating a suboptimal action and underestimating the optimal one up to $\Delta^d_{\widehat{I}^d}/2$.
	\begin{align*}
	&P \left(
	\widehat{I}^d \neq i^*_d \Big| b(d)
	\right)
	=
	P \left(
	\returnestimate{d}{\widehat{I}^d}{b(d)}
	\ge
	\returnestimate{d}{i^*_d}{b(d)}
	\Big| b(d)
	\right)\\
	&\begin{aligned}
	\le
	P \left(
	\returnestimate{d}{\widehat{I}^d}{b(d)}
	\ge
	\return^d_{\widehat{I}^d,b(d)} + \frac{\Delta^d_{\widehat{I}^d}}{2}
	\; \cup \right.\\
	\left. \returnestimate{d}{i^*_d}{b(d)}
	\le
	\return^d_{i^*_d,b(d)} - \frac{\Delta^d_{\widehat{I}^d}}{2}
	\Big| b(d)
	\right)
	\end{aligned}\\
	&\begin{aligned}
	\le \;
	&P \left(
	\returnestimate{d}{\widehat{I}^d}{b(d)}
	\ge
	\return^d_{\widehat{I}^d,b(d)} + \frac{\Delta^d_{\widehat{I}^d}}{2} 
	\Big| b(d)
	\right) + \\
	&P \left(
	\returnestimate{d}{i^*_d}{b(d)}
	\le
	\return^d_{i^*_d,b(d)} - \frac{\Delta^d_{\widehat{I}^d}}{2}
	\Big| b(d)
	\right)
	\end{aligned}
	\end{align*}
	From now on, the proof breaks to the analysis of one of the two terms on the right of the last inequality since both can be considered the same way.
	Let us consider the first term:
	\begin{align*}
	&\quad
	P \left(
	\returnestimate{d}{\widehat{I}^d}{b(d)}
	\ge
	\return^d_{\widehat{I}^d,b(d)} + \frac{\Delta^d_{\widehat{I}^d}}{2} 
	\Big| b(d)
	\right)\\
	&\begin{aligned}
	=
	\sum_{t = 1}^{b(d)}
	&P \left(
	\returnestimate{d}{\widehat{I}^d}{b(d)}
	\ge
	\return^d_{\widehat{I}^d,b(d)} + \frac{\Delta^d_{\widehat{I}^d}}{2}
	\Big| \nbsamples{d}{\widehat{I}^d}{b(d)} = t
	\right) \times\\
	&P \left(
	\nbsamples{d}{\widehat{I}^d}{b(d)}  = t
	\Big| b(d)
	\right)
	\end{aligned}\\
	&\le
	\sum_{t = 1}^{b(d)}
	\exp \left\{
	-\frac{1}{2} (\Delta^d_{\widehat{I}^d})^2 t
	\right\}
	P \left(
	\nbsamples{d}{\widehat{I}^d}{b(d)}  = t
	\Big| b(d)
	\right)\\
	&\le
	\sum_{t = \lceil \rho \ln(b(d)) \rceil}^{b(d)} 
	\exp \left\{
	-\frac{1}{2} (\Delta^d_{\widehat{I}^d})^2 t
	\right\}
	P \left(
	\nbsamples{d}{\widehat{I}^d}{b(d)}  = t
	\Big| b(d)
	\right)\\
	&\le
	\exp \left\{
	-\frac{1}{2} (\Delta^d_{\widehat{I}^d})^2 \lceil \rho \ln(b(d)) \rceil 
	\right\}\\
	&\le
	b(d)^{-\frac{\rho}{2} (\delta^d)^2} 
	\end{align*}
	Where we first write the joint probability, then apply Hoeffding's inequality, followed by Lemma~\ref{lem:lemma1} and the fact that a convex combination is upper bounded by its higher element.
	Similarly to \citet{kocsis2006bandit}, we shall assume that the UCT constant $C_p$ is appropriately chosen for the tail inequalities to be verified.
\end{proof}
\fi

\begin{thm}
	\label{thm:thm1}
	\textbf{Upper bound on the failure probability at depth $d$.}
	For an initial budget of $n$ and for any sub-tree $\mathcal{T}_d$ developed with a budget $b(d)>K$, we have the following recursive relation for the upper bound on the failure probability, conditioned by the initial budget $n$:
	\begin{equation*}
		P(\widehat{I}^d \neq i^*_d | n)
		\le
		\lceil \rho \ln(b(d-1)) \rceil^{-\frac{\rho}{2} (\delta^d)^2}
		.
	\end{equation*} 
	Additionally, for any depth $d \ge 1$ given the initial budget $n$:
	\begin{equation*}
		\begin{cases}
		P(\widehat{I}^d \neq i^*_d | n)
		\le
		{f^{d}(n)}^{-\frac{\rho}{2} (\delta^d)^2}\\
		f: t \mapsto \lceil \rho \ln(t) \rceil 
		\end{cases}
		.
	\end{equation*}
	Where $f^d = f \circ f^{d-1}$ with $f^1 = f$ and $d>0$.
\end{thm}

\if\addproofs1
\begin{proof}
	We write the joint probability distribution:
	\begin{align*}
	P(\widehat{I}^d \neq i^*_d | n)
	&=
	\sum_{t = 1}^{n}
	P(\widehat{I}^d \neq i^*_d | n, b(d) = t)
	P(b(d) = t | n)\\
	&\le
	\sum_{t = 1}^{n}
	t^{-\frac{\rho}{2} (\delta^d)^2}
	P(b(d) = t | n)\\
	&\le
	\sum_{t = \lceil \rho \ln(b(d-1)) \rceil}^{n} 
	t^{-\frac{\rho}{2} (\delta^d)^2}
	P(b(d) = t | n)\\
	&\le
	\lceil \rho \ln(b(d-1)) \rceil^{-\frac{\rho}{2} (\delta^d)^2} 
	\end{align*}
	Where we first applied Lemma~\ref{lem:lemma2} and then used the fact that $b(d) = \nbsamples{d-1}{\widehat{I}^{d-1}}{b(d-1)}$ onto which we apply Lemma~\ref{lem:lemma1}.
	Finally, we use the fact that a convex combination is upper bounded by its higher element.
	The last result comes from the sequence of lower bounds in Lemma~\ref{lem:lemma1}.
\end{proof}
\fi

\begin{figure}
	\includegraphics[width=\linewidth]{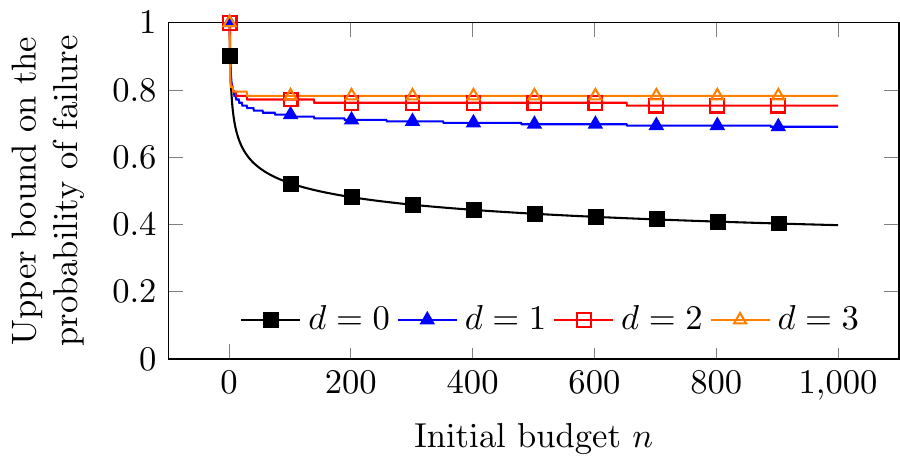}
	\captionof{figure}{
		Upper bound on the probability of failure at depths $d \in \{ 0, 1, 2, 3 \}$ for $C_p = 0.7$ and $\delta^d = 0.27$ for any depth $d$.
	}
	\label{fig:bounds}
\end{figure}

This result shows a logarithmic decay between the upper bounds on the failure probability of two subsequent trees.
Asymptotically, at any depth, this upper bound converges towards zero.
An illustration can be found in Figure~\ref{fig:bounds} for several different depths.
This result highlights the fact that the deeper the sub-tree is, the less one can rely on the recommended action at the root node.
However, we should note that these upper bounds are derived without making further hypotheses on the MDP and express a worst-case value.
Practically, depending on the problem, subsequent sub-trees could be highly relevant \wrt{} the current state of the agent.
We show in the next section that equal performances to \oluct{} can be reached with a smaller computational budget and number of calls to the generative model.

\section{Empirical Analysis}
\label{sec:experiments}

\newcommand\blfootnote[1]{%
	\begingroup
	\renewcommand\thefootnote{}\footnote{#1}%
	\addtocounter{footnote}{-1}%
	\endgroup
}

We compared \oluct{} with \algoname{} on a discrete 1D track environment\blfootnote{
	Code available at:
}\footnote{
	\footnotesize
	\texttt{
		\scriptsize https://github.com/erwanlecarpentier/1dtrack.git
	}
}
and a continuous Physical Travelling Salesman Problem\footnote{
	\footnotesize
	\texttt{
		\scriptsize https://github.com/erwanlecarpentier/flatland.git
	}
}
(PTSP) \citep{perez2012physical}.
We implemented five decision criteria, leading to five variations of \algoname{}.

\subsection{Heuristic decision criteria}

A relevant decision criterion \wrt{} the treated problem allows \algoname{} to discard a sub-tree when its first recommended action may not be state-wise optimal given the current state of the agent.
We implemented five different tests to base this decision on, and evaluated them independently, which led to the following variations of \algoname{}.

\textbf{Plain \algoname{}.} The simplest decision criterion that discards a sub-tree only if its root-node is not fully expanded.

\textbf{State Distribution Modality (SDM-\algoname{}).} Test whether the empirical state distribution is multi-modal or not. If yes, discard the tree if the current state of the agent does not belong to a majority mode.
We define a majority mode by a mode comprising more than $\tau_{SDM} \%$ of the sampled states.

\textbf{State Distribution Variance (SDV-\algoname{}).}
Test whether the empirical state distribution variance is above a certain threshold $\tau_{SDV}$. Discard the tree if it is the case.
For multi-dimensional state spaces such as in the PTSP, the Variance-Mean-Ratio (VMR) is considered for the different orders of magnitude to be comparable.

\textbf{State Distance to State Distribution (SDSD-\algoname{}).}
Compute the Mahalanobis distance \citep{de2000mahalanobis} of the current state from the empirical state distribution. Discard the tree if it is above a selected threshold $\tau_{SDSD}$.

\textbf{Return Distribution Variance (RDV-\algoname{}).} Test whether the empirical return distribution variance is above a certain threshold $\tau_{RDV}$. Discard the tree if it is the case.

A more selective decision criterion can easily be derived by combining the previously described decision criteria and discarding the tree if one of them recommends to do so.

\subsection{1D Track Environment}

\begin{figure}[t]
	\centering
	\includegraphics[
	trim={0cm 0cm 0cm 0cm},
	clip,
	height=1.3cm]{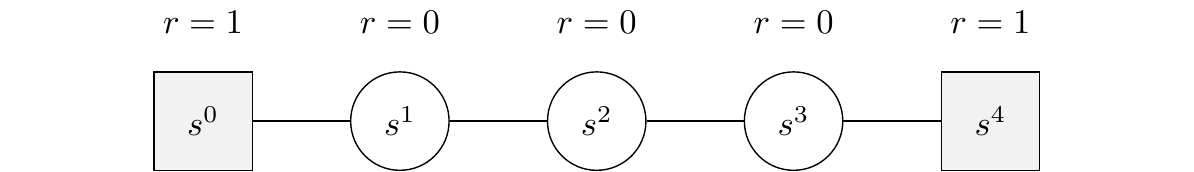}
	\caption{
		1D track environment.
		On top of each cell representing a state is the immediate reward of the transition to this state.
	}
	\label{fig:1dtrack}
\end{figure}

\begin{figure}[]
	\includegraphics[
	trim={0cm 0.4cm 0cm 0.3cm},
	clip,
	width=\linewidth
	]{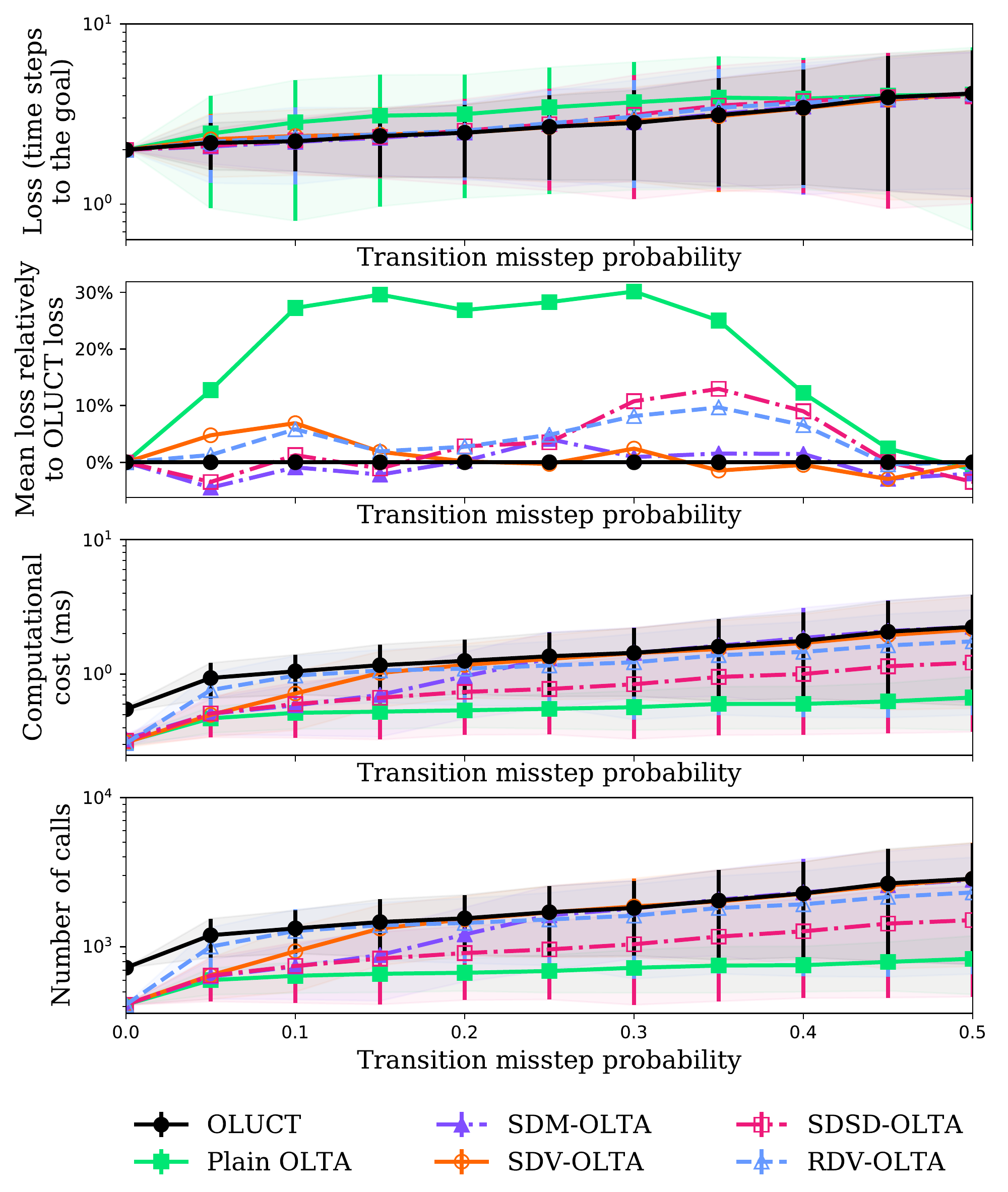}
	\caption{
		Comparison between \oluct{} and \algoname{} on the discrete 1D track environment for varying values of $q$.
	}
	\label{fig:1dtrack_discrete}
\end{figure}

The 1D track environment (Figure~\ref{fig:1dtrack}), is a 1D discrete world where an agent can either go right or left.
The initial state is the ``middle'' state $s_0 = s^2$.
The reward is $0$ everywhere except for the transition to the two terminal states $s^0$ and $s^4$ for which it is $+1$.
The action space is $A = \{ right, left\}$.
We introduce a transition misstep probability $q \in [0,1]$ which is the probability to end up in the opposite state after taking an action, for $i \in \{1, 2, 3\}$: $P(s^{i-1} | s^i, right) = q$ and $P(s^{i+1} | s^i, right) = 1-q$.
The same applies for the $left$ action.
If $q < 0.5$, the optimal policy $\pi_{optimal}$ is to go left at $s^1$, to act randomly at $s^2$ and to go right at $s^3$.
The simulation settings are:
$q \in \{0.0, 0.05, \cdots, 0.5\}$;
$n = 20$ (budget);
$\pi_{default} = \pi_{optimal}$;
$H = 10$ (simulation horizon for $\pi_{default}$);
$C_p = 0.7$;
$\gamma = 0.9$.
The decision criteria parameters were tuned to: $\tau_{SDM} = 80$; $\tau_{SDV} = 0.4$; $\tau_{SDSD} = 1$; $\tau_{RDV} = 0.9$.
We generated $1000$ episodes for each value of $q$ and recorded 3 performance measures:
loss (number of time steps to termination);
computational cost (measured computation time);
and number of calls to the generative model.
We display two different graphs of the loss, the second one  highlights the relative performance between \algoname{} and \oluct{}.

The motivation behind the use of such a benchmark is to test open loop control in a highly stochastic environment where feedback of the current state is highly informative about the optimal action.
First, notice that the parameters are tuned so that the \oluct{} algorithm can easily find the optimal action and that the derived plan at the root node of $\mathcal{T}_0$ is optimal.
In case of misstep for the first action, \algoname{} has to guess that a re-planning should be triggered while \oluct{} does it systematically.
However, the difficulty for \algoname{} is to guess that a misstep occurred and to act accordingly.
As seen on Figure~\ref{fig:1dtrack_discrete}, the non-plain \algoname{} and \oluct{} achieved a very comparable loss.
Plain-\algoname{} had a weaker performance due to its systematic re-use of the sub-trees.
Notice that some variations of \algoname{} such as SDV-\algoname{} achieved a better mean loss than \oluct{} for some values of $q$.
Due to the high variance, this observation cannot lead to the conclusion that \algoname{} can outperform \oluct{}.
However, this emphasizes the fact that the performance are very similar.
In terms of both computational cost and number of calls to the generative model, \algoname{} widely outperforms \oluct{}.
As $q$ increases, this computational gain vanishes and catches up with \oluct{} for SDM-\algoname{} and SDV-\algoname{}.
This accounts for the discriminative power of their decision criteria that discard more trees.
RDV-\algoname{} and SDSD-\algoname{} kept a lower computational cost while reasonably matching the performance of \oluct{}.
Obviously, the computational cost of Plain-\algoname{} stays low.
The apparent similarity between the number of calls to the generative model and the computational cost proves that computing our decision criteria is less expensive than re-planning.

\subsection{Physical Travelling Salesman Problem}

\begin{figure}[t]
	\begin{tabular}{ll}
		\begin{minipage}{3.5cm}
		\includegraphics[
				trim={2cm 2.4cm 2cm 2cm},
				clip,
				height=3.5cm
			]{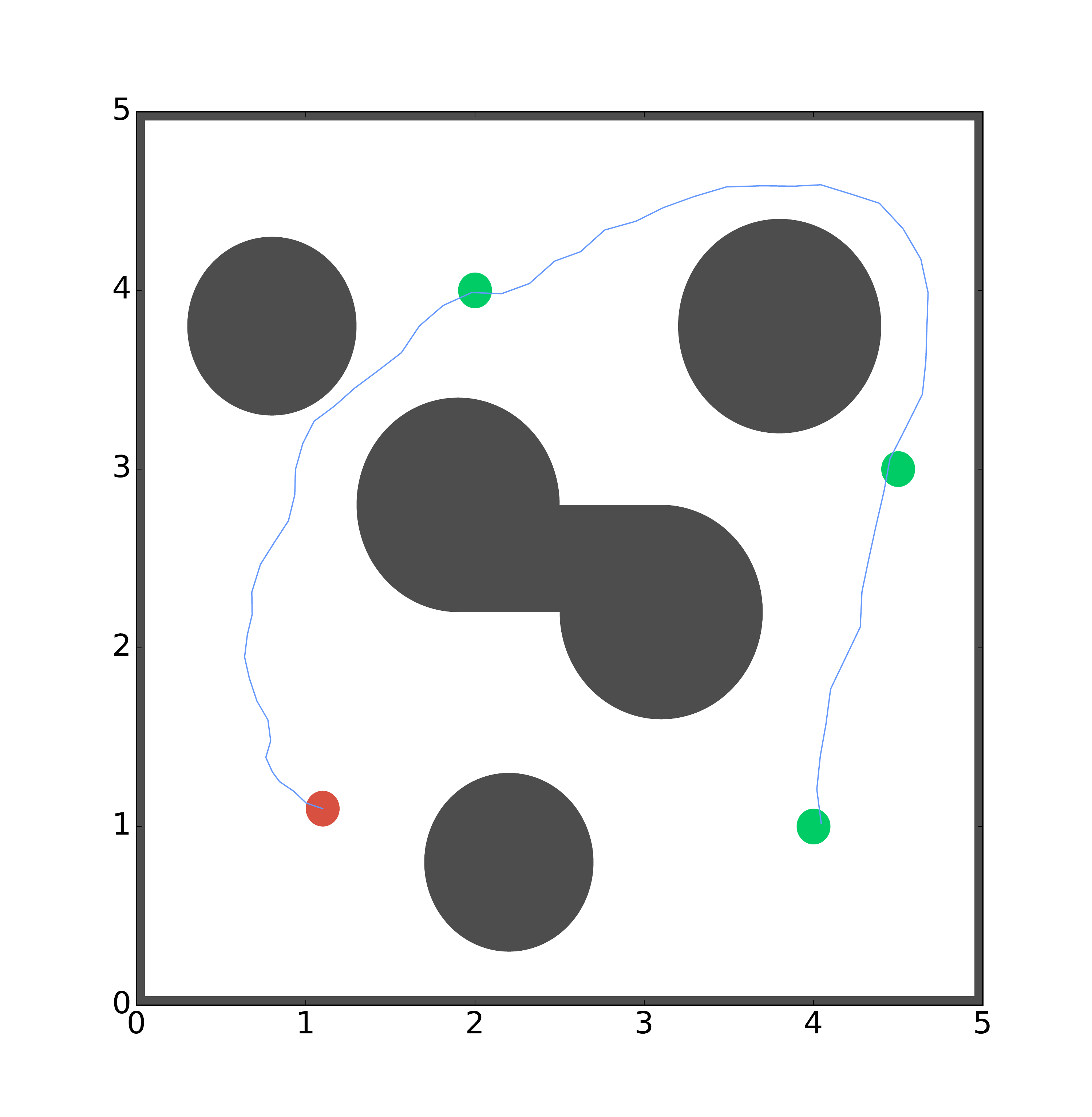}
		\end{minipage}	 &
		\begin{minipage}{4cm}
		Trajectory derived by an \oluct{} algorithm in our PTSP setting. The starting point is displayed in red, the waypoints in green and the walls in grey.
		\end{minipage}
	\end{tabular}
	\caption{
		PTSP illustration
	}
	\label{fig:ptsp_map}
\end{figure}

\begin{figure}[t]
	\includegraphics[
	trim={0cm 0cm 0cm 0cm},
	clip,
	width=\linewidth
	]{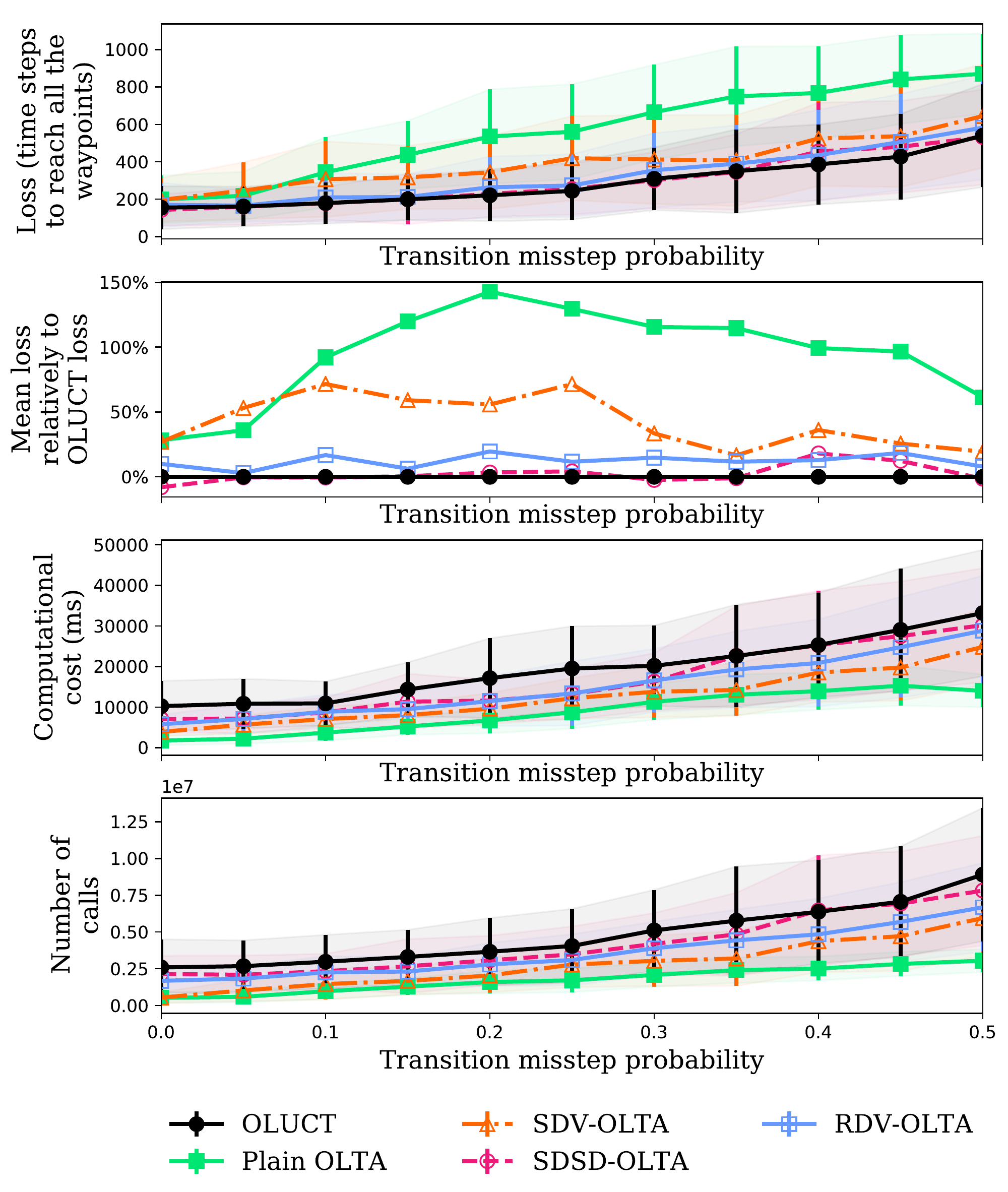}
	\caption{
		Comparison between \oluct{} and \algoname{} on the continuous PTSP for varying values of $q$.
	}
	\label{fig:ptsp_continuous}
\end{figure}

The PTSP is a continuous navigation problem in which an agent must reach all the waypoints within a maze (Figure~\ref{fig:ptsp_map}).
The state of the agent is $s = (x, y, \theta, v) \in \mathbb{R}^4$ \ie{} the 2D position, orientation and velocity.
The action space is $A = \{+d\theta, 0, -d\theta\}$ which consists of the increment, decrement or no-change of the orientation.
The reward is $+1$ when a waypoint is reached for the first time, $-1$ for a wall crash and $0$ otherwise.
The simulation terminates when the agent reaches all the waypoints or a time limit.
The walls cannot be crossed and the orientation is flipped when a crash occurs.
We introduce a misstep probability $q \in [0,1]$ which is the probability for another action to be undertaken instead of the current one.
A Gaussian noise of standard deviation $\sigma_{noise}$ is added to each component of the resulting state from a transition.
The simulation settings are:
$s_0 = (1.1,1.1,0,0.1)$;
$q \in \{0.0, 0.05, \cdots, 0.5\}$;
$\sigma_{noise} = 0.02$;
$n = 300$ (initial tree budget);
$\pi_{default} = \pi_{go-straight}$ that applies no orientation variation;
$H = 50$ (simulation horizon for $\pi_{default}$);
$C_p = 0.7$;
$\gamma = 0.99$.
The provided map is the one depicted in Figure~\ref{fig:ptsp_map} with three waypoints.
The different decision criteria parameters were tuned to:
$\tau_{SDV} = 0.02$;
$\tau_{SDSD} = 1$;
$\tau_{RDV} = 0.1$.
We reserve the development of SDM-\algoname{} in the continuous case for future work.
We generated $100$ episodes for each transition misstep probability and recorded the same performance measures as in the 1D track case.
The results are presented in Figure~\ref{fig:ptsp_continuous}.

\oluct{}, SDSD-\algoname{} and RDV-\algoname{} achieved a comparable loss for every $q$, which shows that our method is applicable to larger scale problems than the 1D track environment.
SDV-\algoname{} reached a lower level of performance.
Plain \algoname{} still realized the highest loss since it is highly sensitive to the stochasticity of the environment.
In terms of both computational cost and number of calls to the generative model, the same trade-off between performance and computational cost is observed.
Plain \algoname{} and SDV-\algoname{} considerably lowered the number of calls at the cost of the performance while SDSD-\algoname{} and RDV-\algoname{} realized a better compromise.
The number of calls to the generative model and the computational cost are quite similar, meaning that --- even with the higher dimensionality of the PTSP compared to the 1D track --- the cost incurred by the decision criteria computation is negligible in comparison to the one incurred by the re-planning procedure.
Notice that SDV-\algoname{} achieved a good cost-performance trade-off in the 1D track environment while not in the PTSP relatively to the other algorithms.
This is explained by the decision criteria's sensitivity to parameter tuning and by the problem-dependent relevance of such a criterion.
For the sake of completeness, we also generated experiments on the continuous 1D track and the discrete PTSP.
The results are available in the Appendix of this paper.
We chose to only illustrate the discrete 1D track and the continuous PTSP for the theoretical interest of the first one and the complexity of the second one.

\section{Conclusion}
\label{sec:conclusion}

We introduced \algoname{}, a new class of tree-search algorithms performing open loop control by re-using subsequent sub-trees of a main tree built with the \oluct{} algorithm.
A decision criterion based on the analysis of
the current sub-tree allows the agent to efficiently determine if the latter can be exploited.
Practically, \algoname{} can achieve the same level of performance as \oluct{} given that the decision criterion is well designed.
Furthermore, the computational cost is strongly lowered by decreasing the number of calls to the generative model.
This saving is the main interest of the approach and can be exploited in two ways: it decreases the energy consumption which is relevant for critical systems with low resources such as Unmanned Vehicles or Satellites; It allows a system to re-allocate the computational effort to other tasks rather than controlling the robot.
We emphasize the fact that this method is generic and can be combined with any other tree-search algorithm than \oluct{}.
Open questions include building non problem-dependent decision criteria, \eg{} by making more restrictive hypothesis on the considered class of MDPs, but also applying the method to other benchmarks and other open loop planners.

\section{Acknowledgements}

This research was supported by the Occitanie region, France.

\bibliographystyle{named}
\bibliography{ijcai_bib}

\newpage
\section*{Appendix}
	
We provide the readers with several additional experiments in similar settings as presented earlier.
The distinction is essentially based on the transition from discrete to continuous state space and vice-versa.
In the paper, we chose to mainly present the two extreme cases of the 1D track and the continuous PTSP for the theoretical interest of the first one and the complexity of the second one.

\section*{Continuous 1D track}

\begin{figure}[h]
	\includegraphics[
	trim={0cm 0cm 0cm 0cm},
	clip,
	width=\linewidth
	]{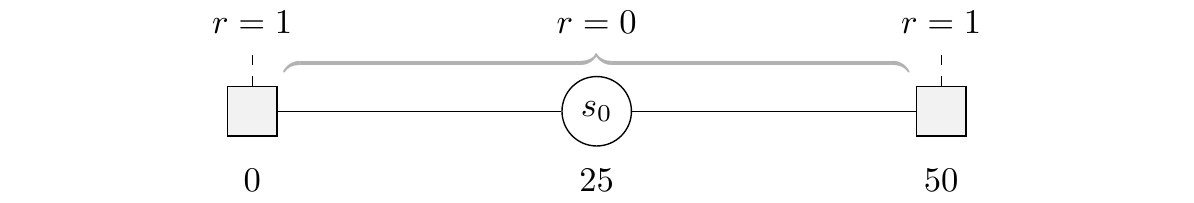}
	\caption{
		Continuous 1D track environment
	}
	\label{fig:1dtrack_continuous_illustration}
\end{figure}

\begin{figure}[t]
	\includegraphics[
	trim={0cm 0cm 0cm 0cm},
	clip,
	width=\linewidth
	]{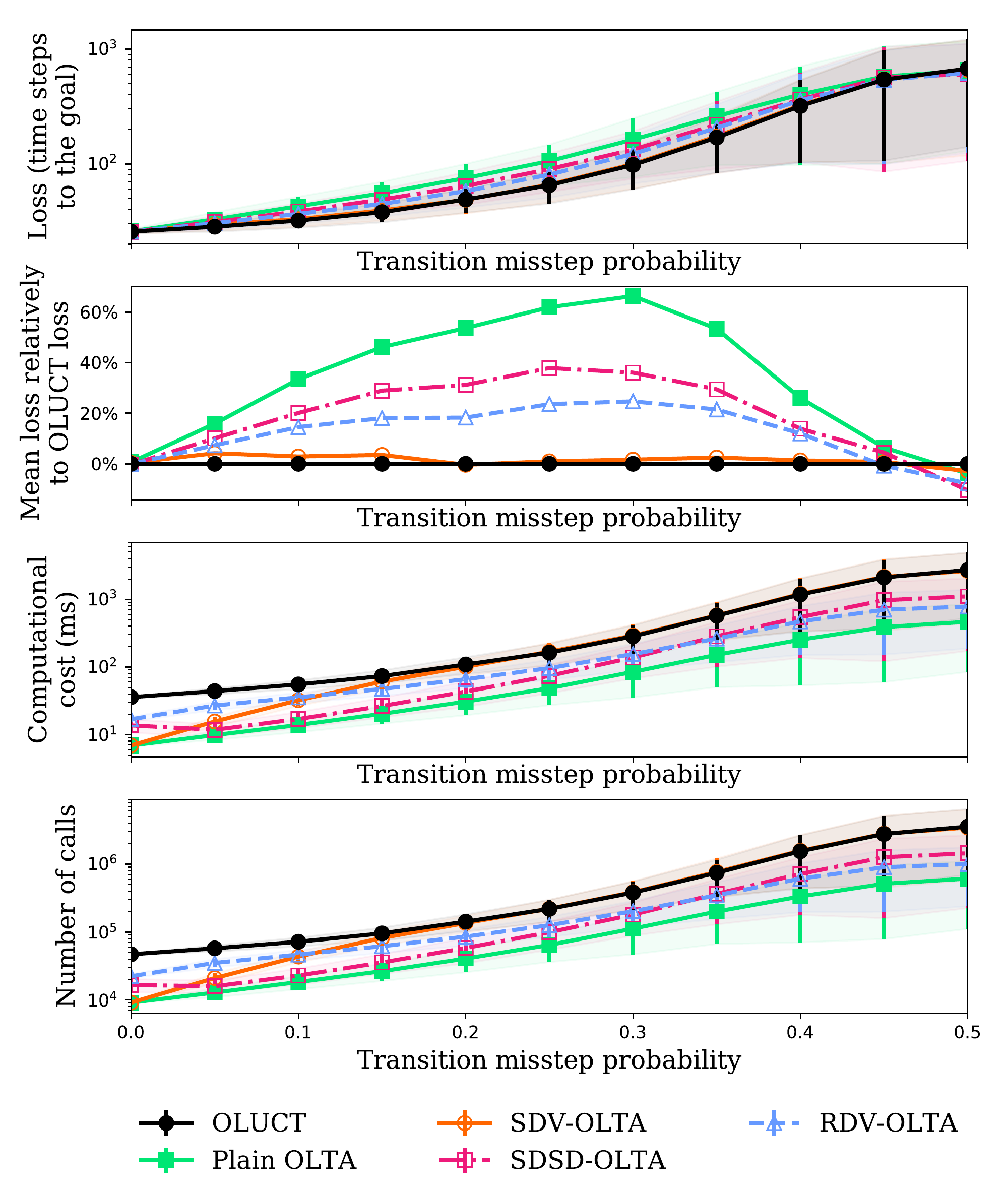}
	\caption{
		Comparison between \oluct{} and \algoname{}{} on the continuous 1D track environment for varying values of $q$.
	}
	\label{fig:1dtrack_continuous}
\end{figure}

In order to test the algorithm on a more complex setting than the discrete 1D track, we extended the latter to the continuous case.
A comprehensive illustration of the environment is provided in Figure~\ref{fig:1dtrack_continuous_illustration}.
The width of the track is $50$, the action space is still $A = \{ right, left\}$ with a magnitude of $1$ for each action.
The agent starts at the middle state $s_0 = 25$.
The reward is zero everywhere except for the two terminal states whose values are $0$ and $50$ where it is $+1$.
To the transition misstep probability presented earlier, we added a Gaussian noise $\epsilon \sim \mathcal{N}(0, \sigma_{noise})$ to the resulting state after each transition.
The simulations are performed with the following settings:
$q \in \{0.0, 0.05, \cdots, 0.5\}$;
$\sigma_{noise} = 0.1$;
$n = 100$ (initial tree budget);
$\pi_{default} = \pi_{optimal}$;
$H = 50$ (simulation horizon for $\pi_{default}$);
$C_p = .7$;
$\gamma = 0.9$;
and the generative model is the true model.
The different decision criteria parameters were selected empirically and set to the following values:
$\tau_{SDV} = 0.4$;
$\tau_{SDSD} = 1$;
$\tau_{RDV} = 5 \cdot 10^{-4}$.
As in the continuous PTSP case, we reserve the development of SDM-\algoname{}{} in the continuous case for future work.
We generated $1000$ episodes for each transition misstep probability.
The results are presented in Figure~\ref{fig:1dtrack_continuous}.
Again, a logarithm is applied for display purposes.
As in the discrete case, \algoname{}{} achieves comparable loss as vanilla \oluct{}.
Particularly, SDV-\algoname{}{} performs as well as \oluct{} on the whole range of misstep probabilities.
In this setting, SDSD-\algoname{}{} and RDV-\algoname{}{} achieved an intermediate loss between Plain-\algoname{}{} and \oluct{}.
In terms of computational cost, two behaviours are observed.
In the case of SDV-\algoname{}{}, the computational gain is relevant for low transition misstep probabilities and catches up with \oluct{} as the latter increases.
This allows the algorithm to achieve the same score as \oluct{}.
In the case of SDSD-\algoname{}{} and RDV-\algoname{}{}, the computational gain seems to be constant on the whole range of transition misstep probabilities.
However, the reached lower performance accounts for the fact that, as for Plain-\algoname{}{}, the decision criteria do not adapt well to the stochasticity increasing, causing the algorithms to discard less trees than needed.
Notice that the computational cost achieved by the SDSD-\algoname{}{} algorithm is greater for $q = 0$ than for $q = 0.05$.
This is due to the fact that its criterion computes the distance between the current state of the agent and the empirical mean of the state distribution normalised by the variance.
This difference comes from the fact that the distribution is mono-modal for $q = 0$ and bi-modal otherwise.
Indeed, in the latter case, the variance increases causing the normalisation to decrease the value of the computed distance.
Additionally, the empirical mean does not correspond to the mean of a mode but a point between the two mode means, which interacts in the opposite way: increasing the computed distance.
In this setting, the interaction of the two mechanisms results in less sub-trees approvals for $q = 0$.
In the discrete case, it does not occur since the current state lies exactly on the mean for $q = 0$ because no Gaussian noise is added to the state transition.
As a result, the distance is always zero.

\section*{Discrete Physical Travelling Salesman Problem}

\begin{figure}[t]
	\centering
	\includegraphics[
	trim={2cm 2cm 2cm 2cm},
	clip,
	height=5.5cm
	]{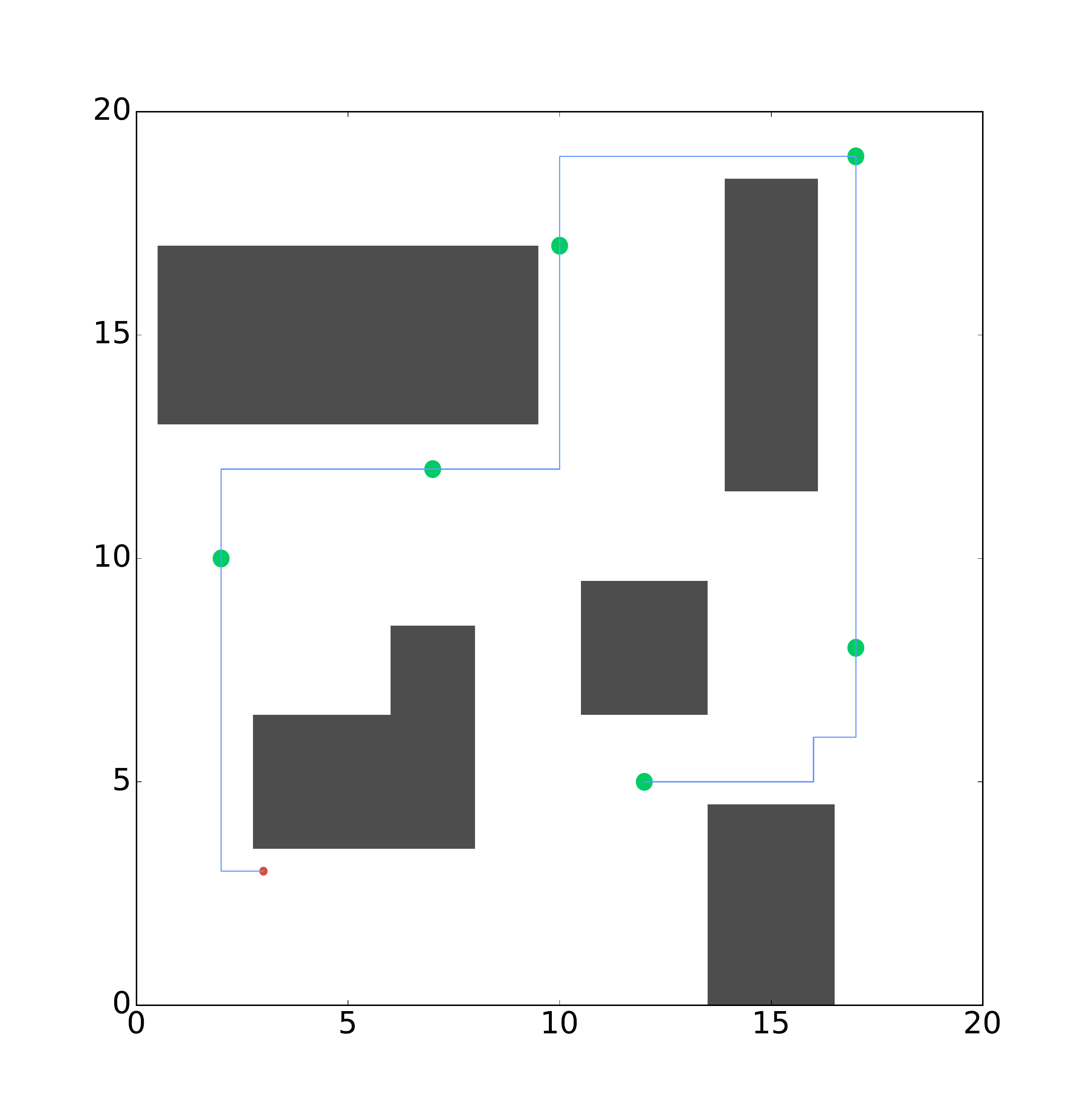}
	\caption{
		Trajectory derived by a \oluct{} algorithm in the discrete PTSP setting. The starting point is displayed in red, the waypoints in green and the walls in grey.
	}
	\label{fig:ptsp_discrete_map}
\end{figure}

\begin{figure}[t]
	\includegraphics[
	trim={0cm 0cm 0cm 0cm},
	clip,
	width=\linewidth
	]{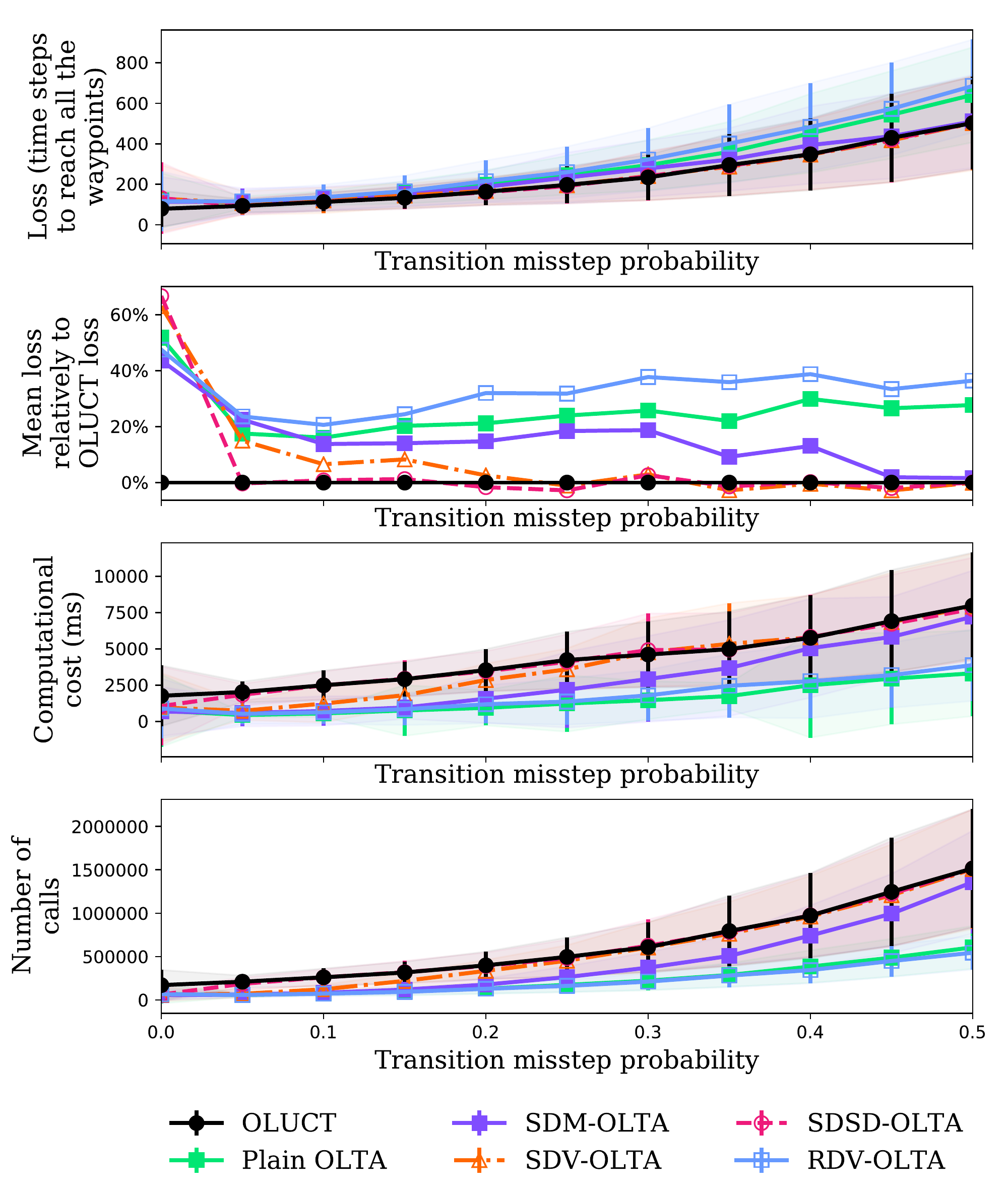}
	\caption{
		Comparison between \oluct{} and \algoname{}{} on the discrete PTSP for varying values of $q$.
	}
	\label{fig:ptsp_discrete}
\end{figure}

We restricted the PTSP to the discrete case.
The resulting problem is a grid-world navigation problem as illustrated in Figure~\ref{fig:ptsp_discrete_map}.
As in the continuous case, the state of the agent is characterized by $s = (x, y, \theta, v) \in \mathbb{R}^4$, respectively the position in the 2D grid-world, the orientation and the velocity.
In our case, we set the velocity to 1 so that the agent only has access to adjacent cells.
The action space is $A = \{ right, down, left, up \}$, each action being the direction of the next adjacent cell reached by the agent.
The reward is set to $+1$ when a waypoint is reached for the first time and to $0$ elsewhere.
We did not penalize the crashes of the agent in the discrete setting because, due to the agility provided by the action space, this would result in being stuck in cells far away from the walls.
We introduce the same misstep probability $q \in [0,1]$ as in the continuous PTSP which is the probability for another action to be undertaken instead of the current one.
The simulations are performed with the following settings:
$s_0 = (3,3,0,1)$;
$q \in \{0.0, 0.05, \cdots, 0.5\}$;
$n = 200$ (initial tree budget);
$\pi_{default} = \pi_{go-straight}$ that applies no orientation variation;
$H = 20$ (simulation horizon for $\pi_{default}$);
$C_p = 0.7$;
$\gamma = 0.99$.
The provided map is the one depicted in Figure~\ref{fig:ptsp_discrete_map} with six waypoints.
The different decision criteria parameters were selected empirically and set as follows:
$\tau_{SDM} = 0.7$;
$\tau_{SDV} = 0.2$;
$\tau_{SDSD} = 1.5$;
$\tau_{RDV} = 0.1$.
Additionally, we provided Plain \algoname{}{} with the ability to discard a sub-tree if the recommended action was not available \ie leading to a wall.
We generated $1000$ episodes for each transition misstep probability.
The results are presented in Figure~\ref{fig:ptsp_discrete}.

As in the continuous case, \algoname{}{} achieves comparable loss as vanilla \oluct{}.
Particularly, SDV-\algoname{}{} and SDSD-\algoname{}{} had a very similar performance on most of the range of misstep probabilities.
SDM-\algoname{}{}, Plain \algoname{}{} and RDV-\algoname{}{} achieved poorer performance but still comparable given the high variance of the losses.
In terms of computational cost, all the variations of \algoname{}{} outperform \oluct{} with an approximately constant gain.
For each one of them, the consequence of this gain was the increasing of the achieved loss, so that each algorithm attained a different compromise between performance and computational cost gain.
	
\end{document}